\documentclass[letterpaper, 10 pt, conference]{ieeeconf}  

\IEEEoverridecommandlockouts                              

\usepackage{amsmath} 
\usepackage{amssymb}  
\usepackage{graphicx}    
\usepackage{gensymb}
\usepackage{subfigure}  
\usepackage{xcolor}
\usepackage{algorithm}      
\usepackage{hyperref}
\hypersetup{
    colorlinks,
    linkcolor={red!50!black},
    citecolor={blue!50!black},
    urlcolor={blue!80!black}
}
\usepackage{booktabs,makecell}
\usepackage{colortbl}
\usepackage{multirow}
\usepackage{algpseudocode}  

\title{\LARGE \bf
Flow with the Force Field: Learning 3D Compliant Flow Matching Policies from Force and Demonstration-Guided Simulation Data}
\author{Tianyu Li$^{1*}$, Yihan Li$^{1*}$, Zizhe Zhang$^{1}$ and Nadia Figueroa$^{1}$%
\thanks{$^{1}$Tianyu Li, Yihan Li, Zizhe Zhang, and Nadia Figueroa are with the GRASP Lab, University of Pennsylvania, PA 19104, USA.}
\thanks{$^{*}$ Equal contribution. Email: \{tianyuli, yianlee\}@seas.upenn.edu}
}
\begin{document}
\maketitle
\thispagestyle{empty}
\pagestyle{empty}

\begin{abstract}

While visuomotor policy has made advancements in recent years, contact-rich tasks still remain a challenge. Robotic manipulation tasks that require continuous contact demand explicit handling of compliance and force. However, most visuomotor policies ignore compliance, overlooking the importance of physical interaction with the real world, often leading to excessive contact forces or fragile behavior under uncertainty. Introducing force information into vision-based imitation learning could help improve awareness of contacts, but could also require a lot of data to perform well. One remedy for data scarcity is to generate data in simulation, yet computationally taxing processes are required to generate data good enough not to suffer from the Sim2Real gap. In this work, we introduce a framework for generating force-informed data in simulation, instantiated by a single human demonstration, and show how coupling with a compliant policy improves the performance of a visuomotor policy learned from synthetic data. We validate our approach on real-robot tasks, including non-prehensile block flipping and a bi-manual object moving, where the learned policy exhibits reliable contact maintenance and adaptation to novel conditions. Project Website: \href{https://flow-with-the-force-field.github.io/webpage/}{flow-with-the-force-field.github.io}.
\end{abstract}

\section{INTRODUCTION}
Contact-rich robotic manipulation tasks demand a delicate balance between precise motion control and compliant force regulation. Mechanical compliance is crucial for successful contact interactions. Visuomotor policies, which are control policy representations that map raw visual observations to motor actions, have emerged as the leading robot learning paradigm for manipulation tasks due to their ease of specification and multi-modal capabilities. Yet, state-of-the-art approaches often ignore compliance and focus only on positional accuracy \cite{chi2023diffusion, zhao2023aloha, zzzral}. Recent advances have highlighted that incorporating compliance or force feedback can drastically improve performance in tasks like object flipping or wiping, where fixed-stiffness controllers fail to handle varying contact conditions \cite{acp, chen2025dexforce}. Some learn variable stiffness profiles from human demonstrations via diffusion models \cite{acp, aburub2024learningdiffusionpoliciesdemonstrations}, while others use reinforcement learning or trajectory optimization to tune compliance \cite{htfl, scape}. 


While these methods achieve strong performance, they typically demand substantial human effort. For instance, ACP \cite{acp} requires hundreds of real demonstrations, and RL or trajectory optimization methods need carefully designed reward functions and extensive data or training. Consequently, current compliance policies lack scalability due to intensive physical data collection or intricate reward engineering. In contrast, our pipeline significantly reduces these burdens by generating force-informed simulation data from a single demonstration, potentially automating compliant policy learning for continuous, contact-rich tasks.

\begin{figure}[!tbp]
  \centering
    \includegraphics[width=0.95\linewidth]{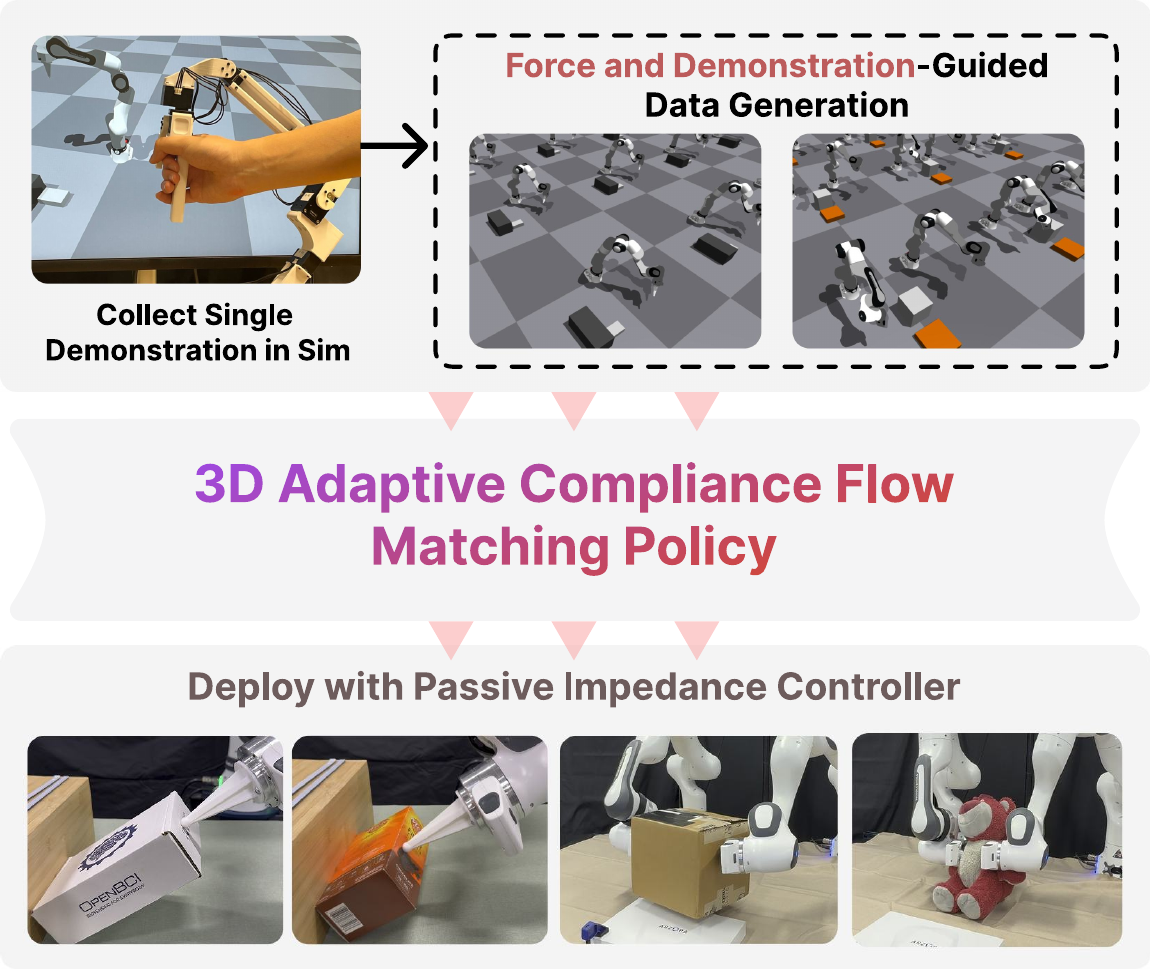}
  \caption{Given \textit{a single demonstration} in simulation, we generate force and demonstration-guided data and transfer to real-world compliant visuomotor policy deployment.
  \label{fig:overview}}
  \vspace{-15pt}
\end{figure}

In this work, we address the challenge of learning vision-force adaptive compliance policies from simulation-generated data instantiated by a single human demonstration. The single demonstration is collected in a simulation environment via teleoperation \cite{liu2025factr}, removing the need for a real robot setup. 
In particular, we propose a lightweight, yet effective, data generation strategy that produces diverse behaviors from a single demonstration in a simulator via force-informed trajectory modulation  \cite{amanhoud2019_ds_force, amanhoud2020_ds_force, acp, chen2025dexforce} and Laplacian editing \cite{nierhoff2016spatial, li2023task, li2025elastic}. These synthetic trajectories are used to train an imitation policy that conditions on 3D pointcloud observations, similar to \cite{Ze2024DP3, chisari2024learning_flow_matching_pointcloud}, end effector pose, and force measurements. The policy outputs task-specific passive impedance parameters that can be executed by a low-level compliance controller on real robots. To learn this complex and multi-modal sensorimotor mapping, we leverage the flow-matching approach \cite{liuflow, zhang2025flowpolicy, black2024pi_0}, allowing high-frequency inference. We introduce using point clouds and force input for the flow matching policy.


While a policy that outputs good trajectories is important, the rollout controller is equally critical for ensuring robustness and safety in the real world, especially under the Sim2Real gap. Thus, we encode the policy rollout poses into a state‑velocity field, which, during execution, is coupled with a Passive Impedance Controller \cite{kronander2015passive} that dampens deviations from the desired velocity. Unlike position‑based controllers that rigidly follow waypoints and often generate abrupt contact forces, this velocity‑based approach reduces energy injection with softer contact, as with state-dependent dynamical system (DS) policies \cite{dsbook}. Combined visuo-force as an input and the compliant vector as the output, this framework reduces the risk of damage due to misalignment while maintaining good performance.

To summarize, we propose a framework for learning adaptive compliant vision‑based policies from simulation data, consisting of three major components, including i) generating force‑informed sim data, ii) policy learning with flow matching, and iii) safe rollout on real hardware with state-velocity fields. Our contributions are fourfold:
\begin{itemize}
    \item Propose a lightweight, yet effective, force‑informed data generation strategy in simulation with virtual targets and Laplacian editing for vision-based imitation learning.
    \item We design an Adaptive Compliant Flow Matching policy that uses point cloud and force as inputs and outputs pose actions and impedance parameters.
    \item We demonstrate zero-shot transfer of our point cloud-based policy to real Franka robots on tasks like box flipping and bimanual grasping, without any real-world demonstrations or sim2real transfer algorithm.
    \item We design a scheme for generating a vector field from policy pose rollouts, enabling passive impedance controller to carry out the compliant policy on real robots with better performance and lower energy injection.
\end{itemize}

\begin{figure*}[!tbp]
    \centering
    \includegraphics[width=0.80\textwidth]{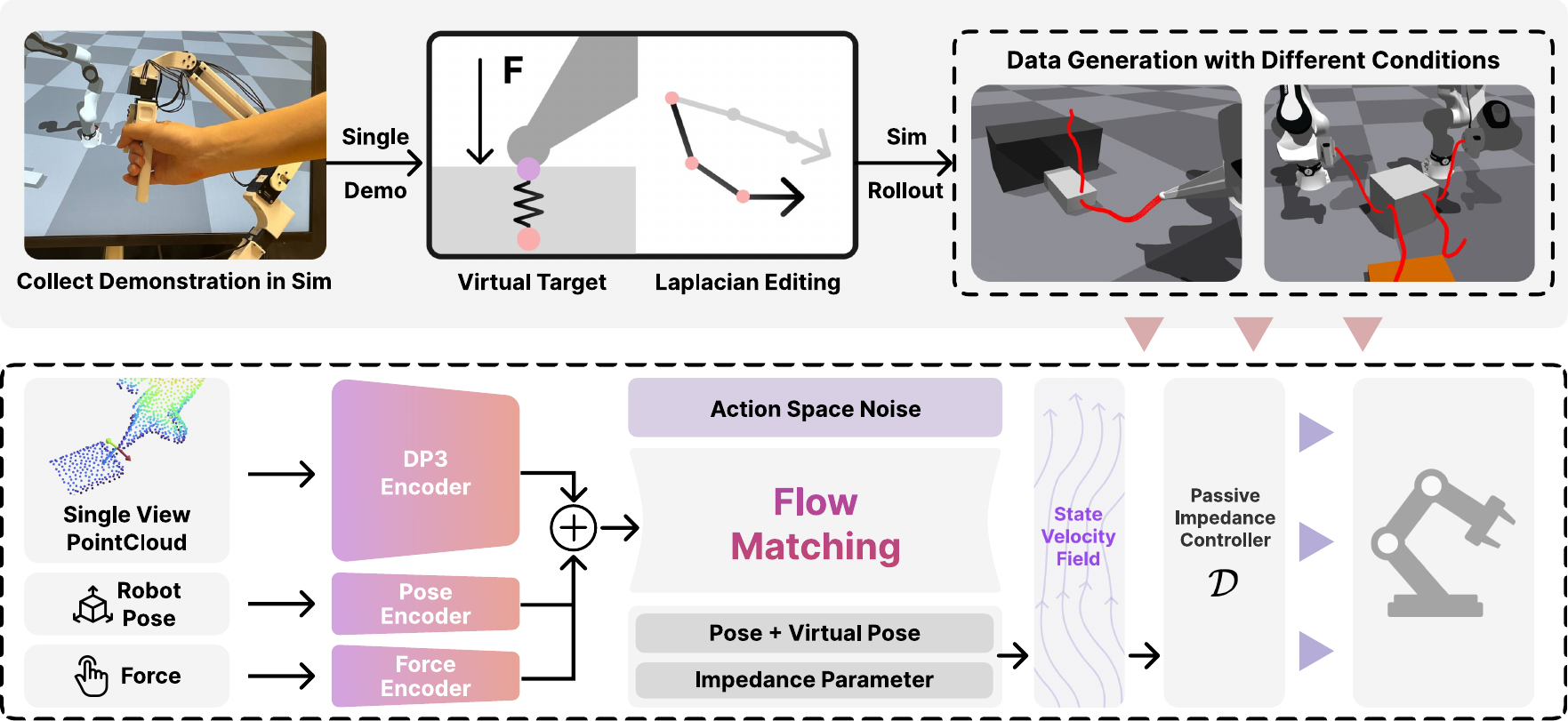}   
    \vspace{-5pt}
    \caption{\textbf{Flow with the Force Field} 3D Compliant Visuomotor Policy Learning Framework: Starting from a single simulation demonstration, we augment data by adding force-informed virtual targets and applying Laplacian editing to generate point cloud and force trajectories beyond the original demo. We train a flow-matching policy that takes point cloud and force as inputs and predicts actions, including an impedance parameter. At rollout, the policy is synthesized into a state-velocity field and executed with a Passive Impedance Controller for compliant behavior. While our data is only generated with one simple box geometry for both tasks, the trained policies produce generalizable capabilities beyond the single shapes using our framework. \label{fig:pipeline}}
    \vspace{-10pt}
\end{figure*}

\section{RELATED WORK}
\subsection{Learning Force-Informed Policies}
Learning policies that integrate visual and force feedback with adaptive compliance has proven highly effective for contact-rich manipulation tasks. \cite{acp} proposed an Adaptive Compliance Policy with a diffusion model trained on human demonstrations, outperforming fixed-stiffness controllers. \cite{chen2025dexforce} introduced DexForce, augmenting kinesthetic demonstrations with force virtual targets to enhance dexterous task performance. FACTR \cite{liu2025factr} combined force-feedback teleoperation with curriculum training, encouraging the model to prioritize joint torque cues rather than solely vision. Our approach differs by employing point clouds instead of RGB inputs, enabling better spatial generalization and robustness to lighting variations. Moreover, rather than extensive real-world demonstrations, we generate synthetic, force-aware data in simulation from a single human demonstration. Reactive methods such as FoAR \cite{he2025foar} and Reactive Diffusion Policy \cite{xue2025reactive_diffusion} integrate vision and force feedback to handle contact-rich tasks, yet still rely heavily on real or teleoperated data. Related approaches include ForceMimic \cite{liu2024forcemimic}, variable impedance learning \cite{martin2019variable}, IRL-based methods for impedance recovery from demonstrations \cite{zhang2021learning}, and RL-based compliance approaches like HTFL \cite{htfl}. These methods generally require extensive real-world demonstrations or involve intricate reward engineering and substantial simulation exploration. In contrast, our method learns adaptive compliance entirely from lightweight, simulation-generated data, eliminating the need for real-world demonstrations or expensive trial-and-error rollouts.
\vspace{-5pt}
\subsection{Data Generation for Sim2Real}
Simulation-based data generation has become critical in reducing real-world demonstration costs. \cite{yang2025physics} introduced PhysicsGen, which generates large-scale datasets via trajectory optimization in simulation from limited human demonstrations, enabling zero-shot sim-to-real deployment. Similarly, \cite{sciencerobotics_tri} uses example-guided RL in simulation and distills policies for whole-body manipulation using state-based inputs and large amounts of simulated interaction. In contrast, our work introduces a direct trajectory warping technique for generating force-informed data for sim2real compliant policy with flow matching \cite{peri2024point, chisari2024learning_flow_matching_pointcloud} on point cloud and force inputs. DemoGen \cite{xue2025demogen} similarly synthesizes novel demonstrations from a single example via TAMP, but focuses on pick-and-place tasks rather than compliant continuous-contact tasks. Approaches such as FORGE \cite{noseworthy2025forge}, which incorporates force information and domain randomization, and DyWA\cite{lyu2025dywa}, with joint impedance in the action space, both use RL to train policies for sim-to-real transfer. However, RL techniques require extra training and configuration, while our method can be directly applied to a human-instantiated demonstration. Frameworks such as \cite{dexpoint} also leverage pointcloud observations and contact reward for sim2real dexterous manipulation, but do not explicitly handle adaptive compliance throughout a task.

In contrast to all aforementioned methods, our framework \textbf{is the first to show} that one can generate force-informed data with a lightweight trajectory modification approach, such as Laplacian Editing \cite{nierhoff2016spatial} and DS force modulation \cite{amanhoud2019_ds_force}, from a single demonstration in simulation. Such data allows training flow-matching policies on point cloud and force inputs, and thus produces compliant policies that transfer zero-shot to real hardware in contact-rich manipulation tasks.

\section{APPROACH}
We assume that a simulation environment (IsaacGym) is instantiated for each target task, following similar data-generation or sim2real setups \cite{yang2025physics, sciencerobotics_tri, lum2025dextrah}, thereby enabling privileged access to information such as object pose and contact forces. The formulation of our framework starts from a single example of human demonstration in simulation $D_h=\left(o^{(i)}_h, a^{(i)}_h\right)_{i=1}^T$ of length $T$.
The observations $o^{(i)}_h = (x_{ee}, F, x_{obj})$ are composed of robot end-effector pose $x_{ee}\in\mathbb{R}^3\times \mathcal{SO}(3)$ and end-effector force measurements $F^r\in\mathbb{R}^3$, as well as object pose $x_{obj}\in\mathbb{R}^3\times \mathcal{SO}(3)$. We define action $a^{(i)}$ as the next end effector pose. We use the privileged information in simulation to help with generating new data. As will be described in Section \ref{sec:data-gen}, $D_h$ is then modified with Laplacian editing and force modulation under domain randomization, which we group the operation as $\mathcal{M}(D_{h})$. This operation generates a more diverse simulation dataset of the task $D_{sim}=\left\{\left(o^{(i)}_s, a^{(i)}\right)\right\}_{i=1}^n$ from $D_h$. 
Note that the $o^{(i)}_s$ in $D_{sim}$ is different from the $o^{(i)}_h$ in $D_{h}$. The human demonstration in simulation $D_{h}$ does not collect point cloud information, while $D_{sim}$ does. $D_{sim}$ is used for training the visuomotor compliant flow matching policy $\pi: \mathcal{O} \rightarrow \mathcal{A}$. The observation space for the policy is $\mathcal{O} = (o_{pc}, o_{ee}, o_f)$ and the action space for the policy is $\mathcal{A} = (a_{ee},d)$, corresponding to the target end-effector pose and compliance gain $d$. During policy rollout on hardware, to improve safety, robustness, and passivity, especially under sim‑to‑real discrepancies, we convert the policy output pose $\mathcal{A} = (a_{ee},d)$ into a state velocity field $\dot{x}_d = f(o_{ee}, d)$. The state velocity field is then fed into a passive impedance controller to produce the desired task space force, which drives the robot, denoted as $F_{ee,d} = D(x) \dot{x}_d$, which damps the deviations from the desired velocity, smoothing motion and suppressing abrupt force spikes; described in Section \ref{sec:compliant}.
Fig. \ref{fig:pipeline} shows a complete view of our pipeline.

\begin{figure}[tbp!]
    \centering
    \begin{minipage}[b]{0.21\textwidth}
        \centering
        \includegraphics[width=\textwidth]{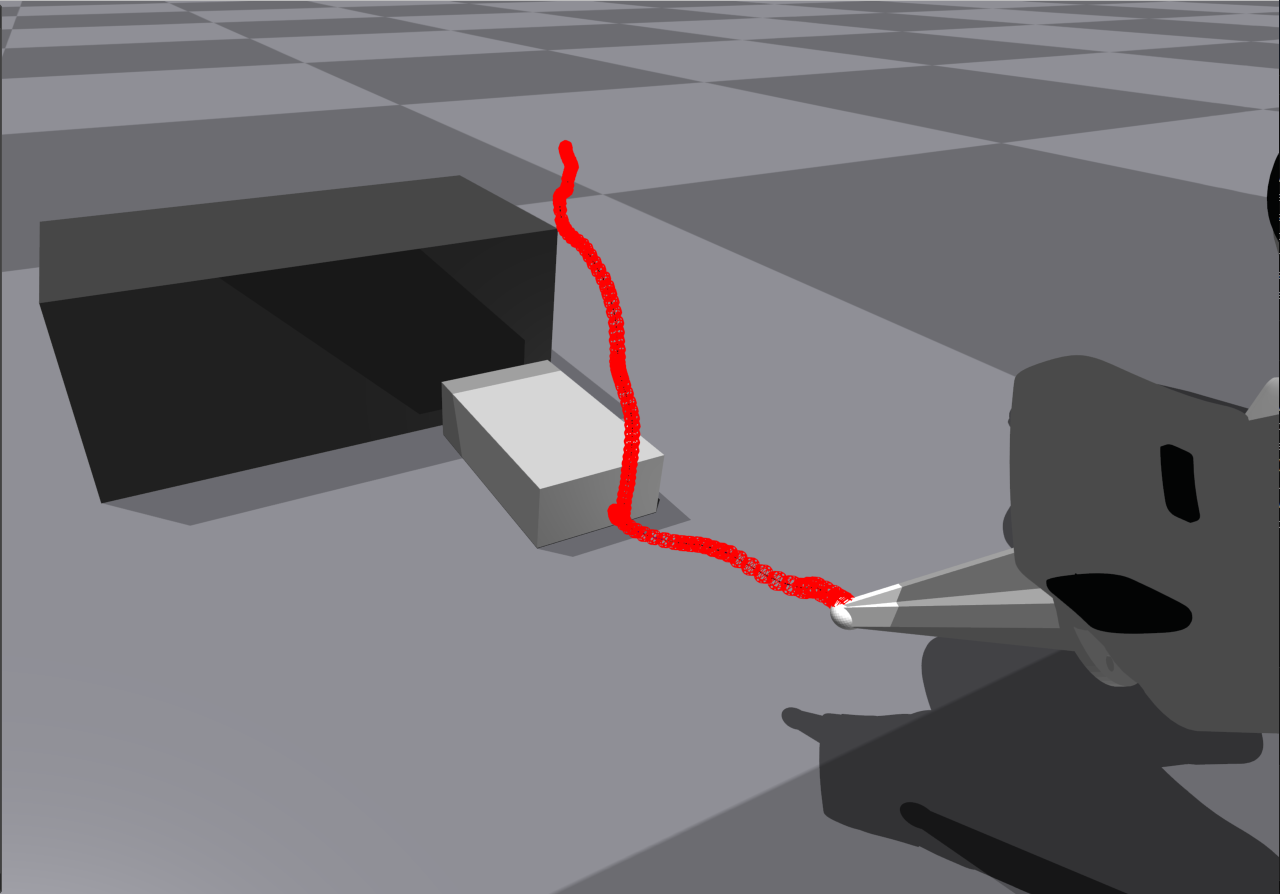}
        \vspace{0.3em}
         {\footnotesize \textbf{(a)} Original reference}
        \label{fig:original_ref}
    \end{minipage}
    \hspace{1pt}
    \begin{minipage}[b]{0.21\textwidth}
        \centering
        \includegraphics[width=\textwidth]{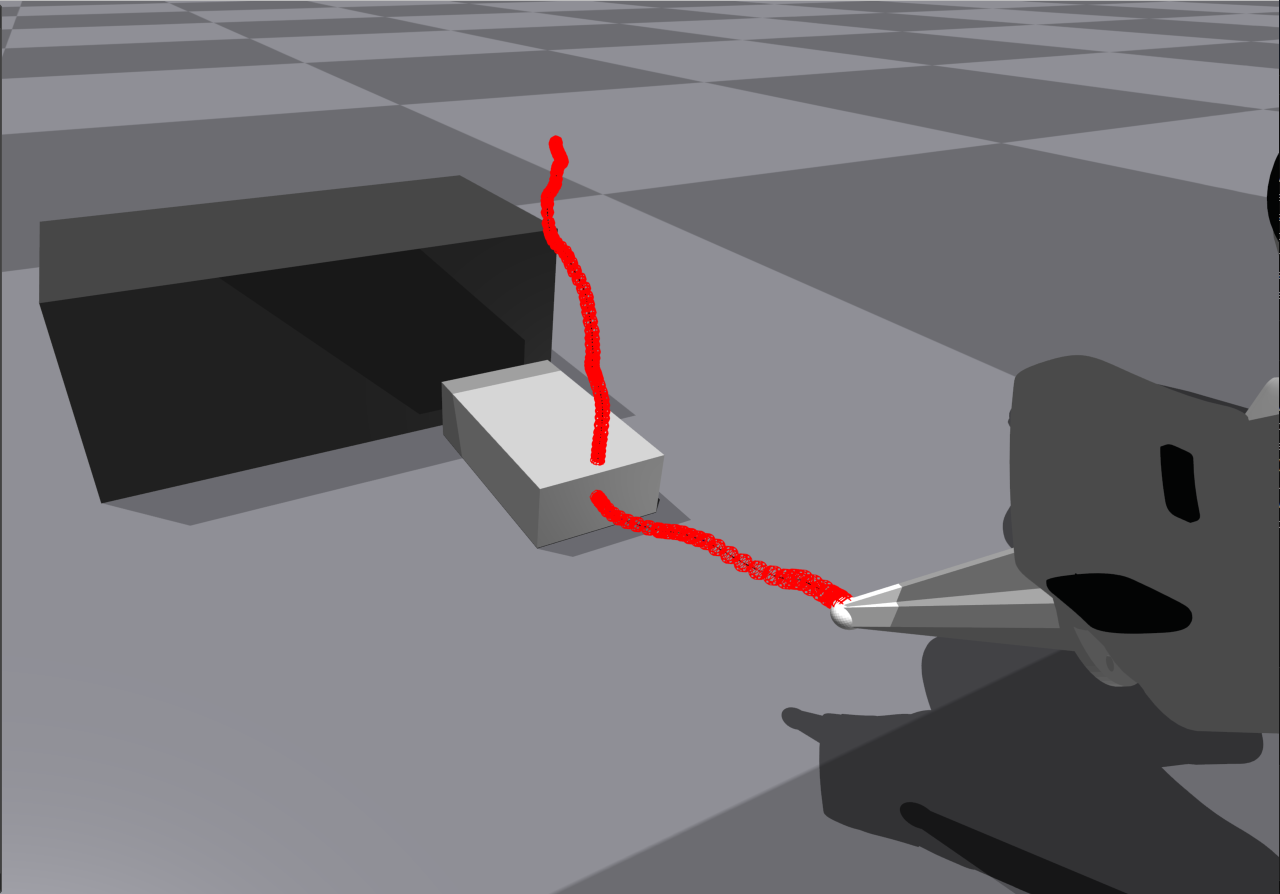}
        \vspace{0.3em}
        {\footnotesize \textbf{(b)} Force-informed} 
        \label{fig:force_modulated}
    \end{minipage}
    
    \begin{minipage}[b]{0.21\textwidth}
        \centering
        \includegraphics[width=\textwidth]{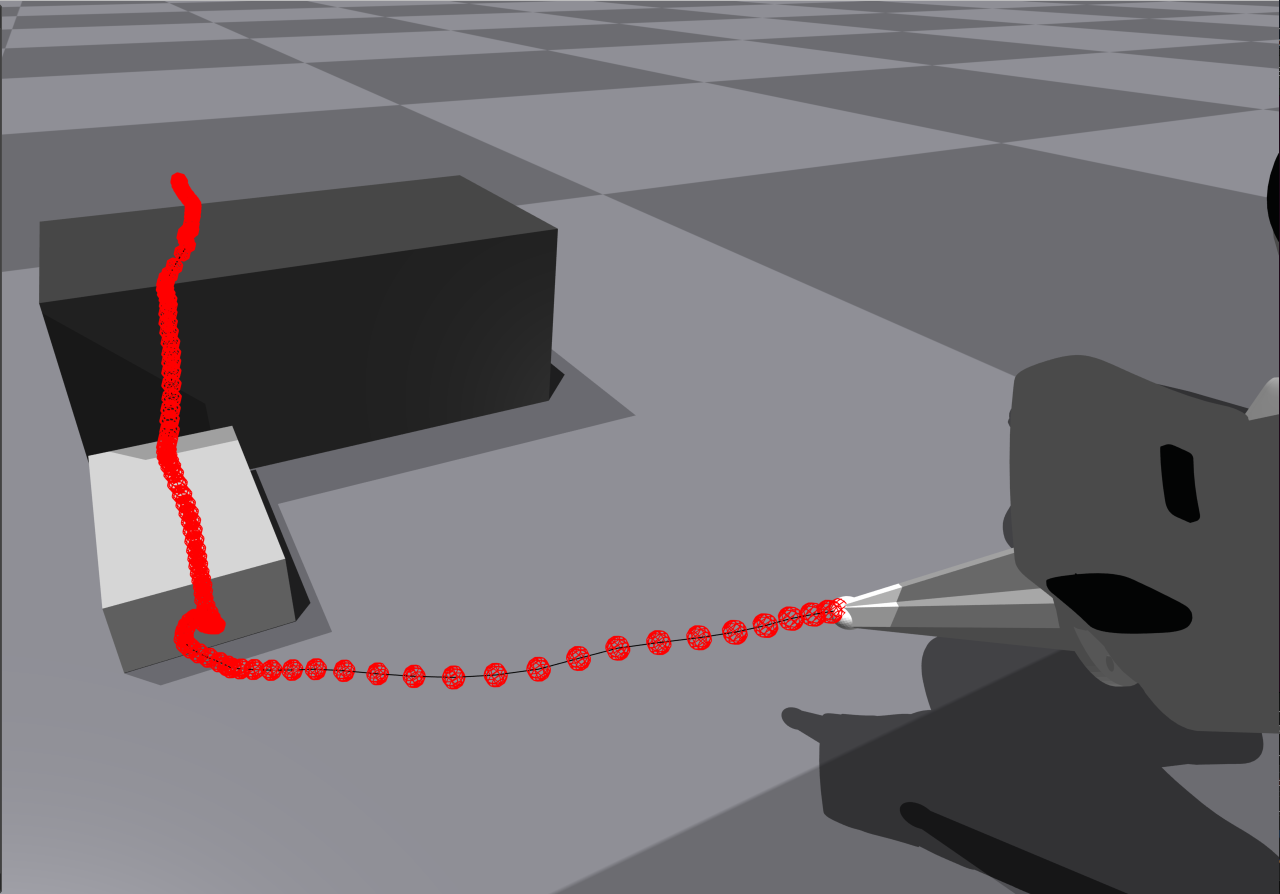}
        \vspace{0.3em}
         {\footnotesize \textbf{(c)} Laplacian editing}
        \label{fig:obj_modulated}
    \end{minipage}%
    \hspace{5pt}
    \begin{minipage}[b]{0.21\textwidth}
        \centering
        \includegraphics[width=\textwidth]{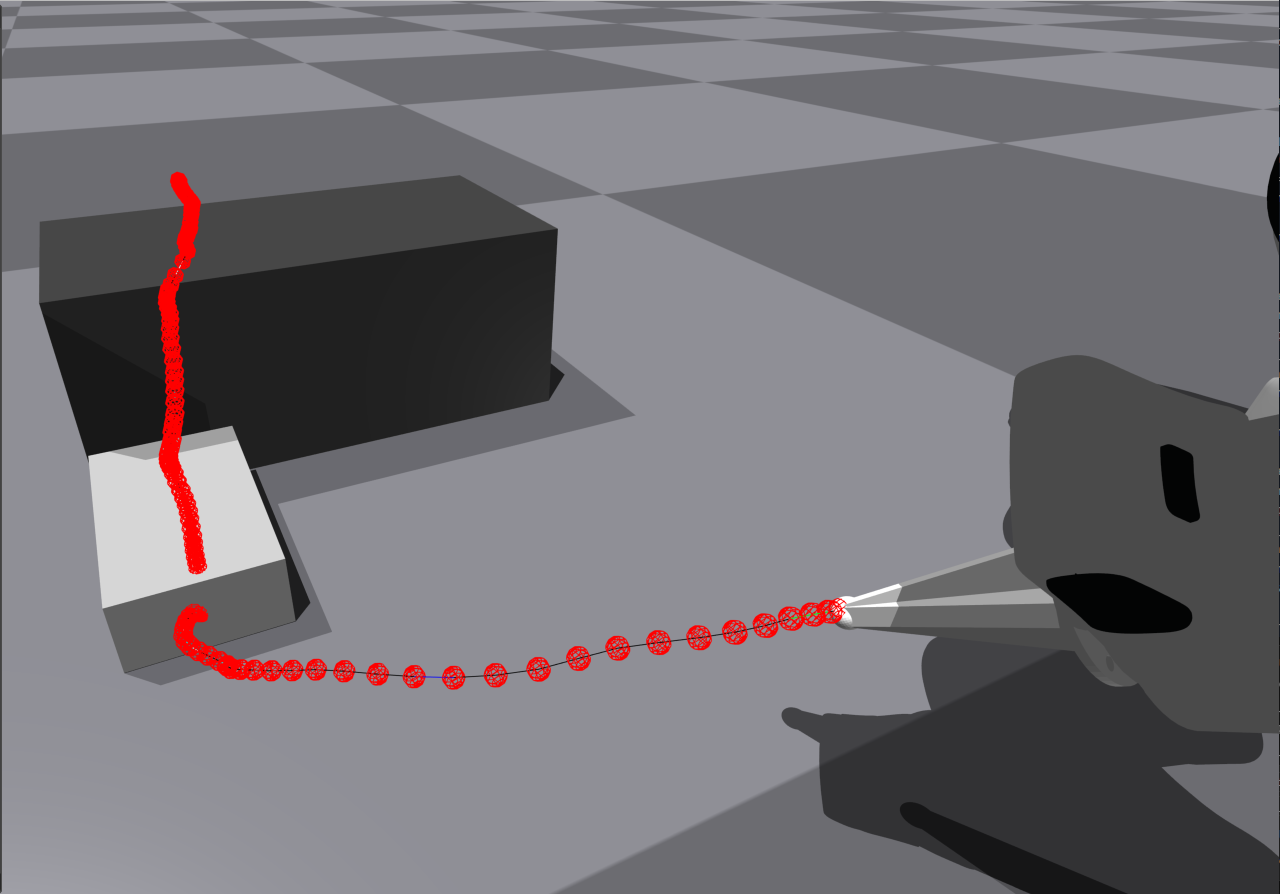}
        \vspace{0.3em}
         {\footnotesize \textbf{(d)} Force-Laplacian}
        \label{fig:object_force_modulated}
    \end{minipage}
    \vspace{-5pt}
    \caption{Example of trajectory warping with different modulation strategies.  (a) is the demonstration trajectory, (b) shows the force-informed trajectory with virtual target, and (c) shows the trajectory warped by object and end effector initial pose, while (d) shows the complete trajectory we use in data generation warped both by object and end effector initial pose and force.\label{fig:datagen_example} }
    \vspace{-10pt}
\end{figure}

\subsection{Generating Force-Informed Data from Simulation}
\label{sec:data-gen}
We provide a light-weight data generation method for force-sensitive tasks without doing any training, but directly warping a single human demonstration under new initial conditions in simulation. As shown in Fig. \ref{fig:pipeline}, we first use FACTR \cite{liu2025factr} with force feedback to teleoperate a Franka in IsaacGym and collect a human demonstration. We then split \(D_h\) into free-space and in-contact segments (Fig.~\ref{fig:datagen_example}): \(D_f\) (length \(T_f\)) where \(\|F\|=0\), and \(D_c\) (length \(T_c\)) where \(\|F\|>0\). Prior to warping the demonstration trajectory, we apply domain randomization to the end effector's pose and the object's pose, mass, and friction in simulation, yielding new initial conditions that differ from those in \(D_h\). Our goal is to generate a new demonstration consistent with these randomized conditions by decomposing and warping \(D_h\).

\textbf{Free-space trajectory:} We warp the transition segment of the end-effector trajectory in \(D_f\) using Laplacian Editing (LE), as LE allows directly modifying an existing trajectory defined by \(m\) constrain waypoints \(\mathbf r \in \mathbb{R}^{d\times m}\) while capturing local properties. We denote the initial end effector trajectory in \(D_f\) as \(\mathbf{X}_{ee}^{f} = [\mathbf{P}_{ee}^{f}, \mathbf{Q}_{ee}^{f}]\), where \(\mathbf{P}_{ee}^{f}\) is the transition part of \(\mathbf{X}_{ee}^{f}\) and \(\mathbf{Q}_{ee}^{f}\) is the rotation part of \(\mathbf{X}_{ee}^{f}\), and the LE-warped trajectoried as \(\mathbf{X}_{ee_w}^{f} = [\mathbf{P}_{ee_w}^{f}, \mathbf{Q}_{ee_w}^{f}]\), in which $\mathbf{X}_{ee_w}^{f}(0)$ is the new initial pose of the end effector. Denoting the new initial pose of the object as $\mathbf{X}_{obj}^{new}(0)=[\mathbf{P}_{obj}^{new}(0), \mathbf{Q}_{obj}^{new}(0)]$, we anchor only two short bands in the LE constraints:
the first \(n_{cs}\) samples at the start and the last \(n_{ce}\) samples at the end
(\(n_{cs}+n_{ce}=m\)). The constraints preserve the \emph{relative} offsets of the first \(n_{cs}\) samples to $\mathbf{P}_{ee_w}^{f}(0)$ and the last \(n_{ce}\) samples to $\mathbf{P}_{obj_w}^{f}(0)$ :
\begin{equation}
\label{eq:LE-start}
{\small
\begin{aligned}
\mathbf{P}_{ee_w}^{f}(t) &= \mathbf{P}_{ee_w}^{f}(0) + \big(\mathbf{P}_{ee}^{f}(t)-\mathbf{P}_{ee}^{f}(0)\big),
&&\hspace{-1em} t=0{:}n_{cs}-1,\\[-3pt]
\mathbf{P}_{ee_w}^{f}(t) &= \mathbf{P}_{obj}^{new}(0) + \big(\mathbf{P}_{ee}^{f}(t)-\mathbf{P}_{ee_w}^{f}(T_f)\big),
&&\hspace{-1em} t=T_f-n_{ce}{:}T_f.
\end{aligned}
}
\end{equation}
Then we first convert the constraint waypoints \(\mathbf r\) in Cartesian space into Laplacian coordinates \(\Delta\) with the graph Laplacian matrix \(L \in \mathbb{R}^{m\times m}\)\cite{nierhoff2016spatial},

\begin{equation}
\label{eq:Laplacian}
{\small
L_{ij} = 
\begin{cases}
1, & i=j,\\[-2pt]
-\dfrac{w_{ij}}{\sum_{j\in\mathcal N_i} w_{ij}}, & j\in \mathcal N_i,\\[-2pt]
0, & \text{otherwise}.
\end{cases}
}
\end{equation}
where \(\mathcal N_i\) are a set of neighbor points \(\mathbf r_j\) for waypoint \(\mathbf r_i\), and \(w_{ij}\) is a weight set to \(1\) for this work. One can obtain \(\Delta = L\mathbf r\), where \(\Delta\) is a concatenation of
the Laplacian coordinate for each waypoint
\begin{equation}
\boldsymbol{\delta}_i
= \sum_{j\in \mathcal N_i}
\frac{w_{ij}}{\sum_{j\in \mathcal N_i} w_{ij}}\,
\bigl(\mathbf r_i - \mathbf r_j\bigr).
\end{equation}
The matrix \(L\) can be singular, so one can impose constraints on the system \(L\mathbf r = \Delta\) when solving for new waypoints \(\mathbf r\) to achieve editing \cite{nierhoff2016spatial}. To warp the rotational components of the free-space trajectory, we apply SLERP (Spherical Linear Interpolation of Rotations) between $\mathbf{Q}_{ee_w}^{f}(0)$ and $\mathbf{Q}_{ee_w}^{c}(0)$, interpolating the rotation part to the same length of $T_f$ which yields a smooth rotational transition over the entire free-space segment, where $\mathbf{Q}_{ee_w}^{c}(0)$ is the rotation part of the first data point in the end effector trajectory of \(D_c\) and will be given in the next section. Combining these orientations with the warped positions gives the complete modulated trajectory $\mathbf{X}_{ee_{w}}^{f}\in\mathbb{R}^3\times \mathcal{SO}(3)$. As illustrated in Fig.~\ref{fig:datagen_example}(a)–(c), the object-pose modulation stretches the free-space path and smoothly reorients it in accordance with the object’s pose.

\textbf{In-contact trajectory:} For the in-contact part, we maintain an object-centric frame for trajectory warping. Let \(\mathbf{X}_{ee}^{c} = [\mathbf{P}_{ee}^{c}, \mathbf{Q}_{ee}^{c}]\text{, and }\mathbf{X}_{obj}^c=[\mathbf{P}_{obj}^c, \mathbf{Q}_{obj}^c]\) denote the in-contact end effector and object pose trajectories in the demonstration. For every data point, the warped end-effector pose $\mathbf{X}_{ee_{w}}^{c}(t)=[\mathbf{P}_{ee_{w}}^{c}(t),\mathbf{Q}_{ee_{w}}^{c}(t)]$ and force $F_{w}(t)$, $t = T_f+1, ..., T_f+T_c$ are given by:
\begin{align}
\mathbf{P}_{ee_{w}}^{c}(t) &= \mathbf{P}_{obj}^{new}(0) + \Delta\mathbf{Q}_{obj}(t)\,(\mathbf{P}_{ee}^{c}(t)-\mathbf{P}_{obj}^c(t)) \label{eq:ee_p_warp}\\
\mathbf{Q}_{ee_{w}}^{c}(t) &= \Delta\mathbf{Q}_{obj}(t)\,\mathbf{Q}_{ee}^{c}(t)\label{eq:ee_q_warp}\\
F_{w}(t) &= \Delta\mathbf{Q}_{obj}(t)\,F(t)\label{eq:force_warp}
\end{align}

Where $\Delta\mathbf{Q}_{obj}(t) = \mathbf{Q}_{obj}^{new}(0)\,{\mathbf{Q}_{obj}^c(t)}^{-1}$ . Equation~\eqref{eq:ee_p_warp}\eqref{eq:ee_q_warp} preserves the relative end-effector pose in the object frame while Equation~\eqref{eq:force_warp} rotates the demonstrated force into the randomized object frame. Inspired by \cite{amanhoud2019_ds_force, amanhoud2020_ds_force}, which use an end effector velocity orthogonal to the contact surface to do force modulation, and \cite{acp, chen2025dexforce},  which steer toward a virtual target in the measured force direction, we construct a force-informed virtual target $x_{ee}^v$ for each data point $\mathbf{x}_{ee_{w}}^{c}$ in $\mathbf{X}_{ee_{w}}^{c}$ as:
\begin{equation}
    \mathbf{X}_{ee_{w}}^{cv}(t) = \mathbf{X}_{ee_{w}}^{c}(t) + k_fF_w(t)
\end{equation}
where $k_f$ is a hand-tuned parameter controlling the modulation strength. We claim that without the force-informed virtual target tracking, the task is not able to succeed because it has no intention of maintaining continuous contact. An analogy for the virtual target is a target pose inside the object. Moving to that target will generate force on the object surface. Comparing Fig. \ref{eq:damping-schedule}(a) and \ref{eq:damping-schedule}(b), we could see that the force-informed trajectory gets into the object surface after adding the virtual target. In our method, the compliance scheduling term $d$ is supervised by force feedback: during policy learning, the contact force measured in simulation is converted to a damping target through a fixed linear mapping, and the flow-matching policy is trained to predict $d$ to match this target, which will be introduced in Section \ref{sec:compliant}.


\subsection{Compliant 3D Visuomotor Flow Matching Policy}
\label{sec:3D-flow-matching}
To learn visuomotor policies from simulation-generated data, we use a flow matching framework \cite{liuflow, chisari2024learning_flow_matching_pointcloud, zhang2025flowpolicy} as our core method. Flow matching runs at a higher frequency than diffusion-based models, making it suitable for contact-rich tasks. Prior work has shown that pointcloud inputs significantly enhance data efficiency and spatial generalization \cite{Ze2024DP3, xue2025demogen}. Accordingly, we adopt a single-view point cloud as input, eliminating reliance on privileged object pose information. Thus, our observation space is defined as $\mathcal{O} = (o_{pc}, o_{ee}, o_f)$.

We use DP3 Encoder from \cite{Ze2024DP3} to encode the point cloud $o_{pc}$. We also follow the same practice for processing the data, including furthest point sampling to 1024 points as well as cropping for the workspace. To be more contact-aware, we also include force $o_{f}$ as part of the observation, which we use a simple 3-layer MLP to output a 64-D feature vector. The input also includes a feature vector encoded by a 3-layer MLP for the robot pose $o_{ee}$. We then concatenate the three feature vectors as the observation condition. 

The flow matching backbone in Fig. \ref{fig:pipeline}, a conditional U-Net same as \cite{Ze2024DP3}, but with a different output space. In order to achieve compliant motion with contact in the environment, we augment the flow matching action space with a virtual pose trajectory $X_{vt}^{t...t+H}$ and an impedance parameter trajectory $d^{t...t+H}$. Therefore, the policy output is $A = (X_{ref}^{t...t+H}, X_{vt}^{t...t+H}, d^{t...t+H})$, which are the robot end effector reference pose trajectory $X_{ref}^{t...t+H}$, virtual pose trajectory $X_{vt}^{t...t+H}$, as well as impedance parameter trajectory $d^{t...t+H}$. We use $H=16$ as the policy output horizon in this work.
We will describe the reference pose, virtual pose, and the impedance parameter in detail in Section \ref{sec:compliant}. Next, we formally define the flow matching component. 

The goal of conditional flow matching is to estimate a vector field $\boldsymbol{v_{\theta}}: \mathbb{R}^d \times[0,1] \rightarrow \mathbb{R}^d$, such that integrating the Ordinary Differential Equation (ODE)
\begin{equation}
    \frac{\mathrm{d} z_t}{\mathrm{d} t}=\boldsymbol{v_{\theta}}\left(z_t, t\right),
\end{equation}
on time $t \in[0,1]$ transports $z_0$ from base distribution $p_0$ to the target distribution $z_1$ from $p_1$. The base distribution $p_0$ is typically chosen as Gaussian noise $\mathcal{N}(0, I)$. To learn such a flow, training proceeds by matching $\boldsymbol{v_{\theta}}$ along an interpolation path $z_t=t z_1+(1-t) z_0$. By taking the derivative, one will obtain the direction of a linear path $\boldsymbol{u}=z_1 - z_0$, which is the target velocity. We parametrize the $\boldsymbol{v_{\theta}}$ with a conditional U-Net and  matching the $\boldsymbol{u}$ and $\boldsymbol{v_{\theta}}$ by solving the following regression problem \cite{chisari2024learning_flow_matching_pointcloud},
\begin{equation}
\min _{\boldsymbol{v}} \mathbb{E}_{z_0 \sim p_0, z_1 \sim p_1}\left[\int_0^1\left\|\boldsymbol{u}\left(z_1, z_0\right)-\boldsymbol{v}_\theta\left(z_t, t\right)\right\|_2^2 \mathrm{~d} t\right].
\end{equation}

In this work, the target sample $z_1$ is the action in trajectory data $D_{sim}$ generated from the simulation. During inference, we first initiate a noisy action sample $A_0$ from $p_0$. Then the ODE is integrated with Euler from $t=0$ to $t=1$ conditioned on the observation $\mathcal{O}$ includes pointcloud $o_{pc}$, end-effector pose $o_{ee}$, and end-effector force $o_f$,
\begin{equation}
    A_{t + \delta} = A_{t} + \delta \boldsymbol{v_{\theta}}(A_{t}, \mathcal{O}),
\end{equation}
which we set the $\delta=0.1$ for all of our experiments. After obtaining the actions rollout $A = (X_{ref}^{t...t+H}, X_{vt}^{t...t+H}, d^{t...t+H})$, rather than tracking this trajectory with classical position-based impedance controller as commonly done in prior works, we introduce a different compliant rollout technique in the next section.

\subsection{Compliant Policy Rollout with State Vector Field}
\label{sec:compliant}
After obtaining the policy rollout, we execute the resulting trajectory using a passive impedance controller in task space \cite{kronander2015passive}. This controller ensures passive interaction with the environment, which is crucial when contact leads to constraints or sticking. Although the impedance parameter has been referenced as part of the policy action, we now formalize it. We begin with the control law from \cite{kronander2015passive}:
\begin{equation}
F_c = G_x(x) - D(x)(\dot{x} - f(x)),
\label{eq:passive}
\end{equation}
where $G_x(x)$ is the gravity compensation term (omitted from now on), $D(x) \in \mathbb{R}^{d \times d}$ is the damping matrix, $\dot{x} \in \mathbb{R}^d$ is the current end-effector velocity, and $f(x) = \dot{x}_d$ is the desired velocity from the policy. The damping matrix $D(x)$ is structured via an eigendecomposition:
$D(x) = V(x) \Lambda V(x)^\top$,
where $\Lambda$ is a diagonal matrix of damping gains and $V(x) = [v_1, v_2, v_3]$ is an orthonormal matrix of eigenvectors, with $v_1$ aligned with the desired direction $f(x)$.

While damping-only impedance has the advantage of passivity, reducing damping sacrifices trajectory tracking, potentially causing distribution shifts. Moreover, tuning directional damping gains often requires significant manual effort and may degrade performance. To avoid these, we fix the damping magnitude and instead modulate the desired velocity direction to shape compliance. Unlike the flow matching vector field, this vector field is in the end effector task space during the execution stage. Specifically, we consider only the desired velocity component from Equation \ref{eq:passive}:
\begin{equation}
F_d = D(x) f(x).
\label{eq:simple_passive}
\end{equation}
where $f(x)$ is collinear with $v_1$, which we can simplify to $F_d = v_1 f(x)$. We then decompose the desired velocity into magnitude and direction:
$f(x) = k \hat{u}$,
where $k = 0.1$ is the desired velocity magnitude, which can be adjusted online or designed, and $\hat{u} \in \mathbb{R}^d$ is the direction vector, which is your current desired moving direction. 

Let $x_{curr} \in R^N$ be the robot's current end effector position, the rollout reference trajectory from flow matching be $X_{ref}^{t...t+H}$, and the virtual target trajectory be $X_{vt}^{t...t+H}$. Each reference $X_{ref}^{t}$ has one corresponding virtual target $X_{vt}^{t}$, which $X_{vt}^{t} - X_{ref}^{t}$ provides the compliant direction at $X_{ref}^{t}$. Therefore, we now define $\hat{t} = X_{ref}^{t+1} - X_{ref}^{t}$ and the compliant direction $\hat{n} = X_{vt}^{t} - x_{curr}$. In ideal case, $x_{curr} \approx X_{ref}^{t}$.  We construct $\hat{u}$ to blend between the reference motion direction $\hat{t}$ and compliant direction $\hat{n}$:
\begin{equation}
\hat{u} = \frac{d \hat{n} + \hat{t}}{|| d \hat{n} + \hat{t} ||},
\end{equation}
where $d \in \mathbb{R}_{>0}$ is a gain controlling convergence toward the virtual target and the degree of compliance. This gain is the impedance parameter of our flow-matching policy. During training, the gain $d$ is scheduled based on the estimated force magnitude $|\mathcal{F}|$ collected by the force sensor in the simulation. Similar to the stiffness in \cite{acp}. We define it as:
\begin{equation}
\label{eq:damping-schedule}
\resizebox{0.75\hsize}{!}{%
$d = \begin{cases}
d_{\uparrow}, & |\mathcal{F}| < \mathcal{F}{\downarrow} \\
d{\uparrow} - (d_{\uparrow} - d_{\downarrow}) \frac{|\mathcal{F}| - \mathcal{F}{\downarrow}}{\mathcal{F}{\uparrow} - \mathcal{F}{\downarrow}}, & \mathcal{F}{\downarrow} \leq |\mathcal{F}| \leq \mathcal{F}{\uparrow} \\
d{\downarrow}, & |\mathcal{F}| > \mathcal{F}{\uparrow}
\end{cases}$%
}
\end{equation}
where ${\uparrow}$ and ${\downarrow}$ are hardware-specific maximum and minimum values. This formulation enables directional compliance to emerge naturally from modulating $f(x)$, without sacrificing tracking fidelity. While this formulation might break passivity with such usage, we argue that this state-dependent vector field rollout will allow safer interruption during execution when interacting with the environment. Furthermore, we will show in the experiments that this formulation actually offers better performance in the real world contact tasks while at the same time injecting less energy.

\section{EXPERIMENT}
\begin{table*}[!th]
   \centering
   \hspace{2pt}
  \begin{minipage}{.45\linewidth}
    \centering
    \caption{Domain Randomization Testing Range \label{tab:randomization}}
    \vspace{-5pt}
    \begin{tabular}{ccc}
    \toprule
    & \textbf{Block Flipping} & \textbf{Bi-manual Moving} \\
    \hline
    Robot End Effector [cm] & $[\pm15,\,\pm5,\,\pm3]$ & $[\pm15,\,\pm5,\,0.0]$\\
    Object Translation [cm] & $[\pm15,\,\pm5,\,0.0]$ & $[\pm10,\,\pm15,\,0.0]$\\
    Object Orientation [$\degree$] & $[0.0,\,0.0,\,0.0]$ & $[0.0,\,0.0,\,\pm20]$ \\
    Object Mass [kg] & $[0.1,\,0.8]$ & $[0.1,\,0.8]$ \\
    Friction & $[0.2,\,1.0]$ & $[0.2,\,1.0]$\\ 
    \hline
    \rowcolor{gray!7} \textbf{Success Rate [\%]} & $97.6$ & $82.9$\\ 
    \bottomrule
    \end{tabular}
    \vspace{-5pt}
  \end{minipage}%
  \hspace{30pt}
  \begin{minipage}{.465\linewidth}
  \vspace{-10pt}
    \centering
    \caption{Success Rates for Spatial and Object Generalization \label{tab:success_rate}}
    \vspace{-5pt}
    \begin{tabular}{ccccc}
    \toprule
    \multirow{2}{*}{\parbox{1cm}{Method}} & \multicolumn{2}{c}{\textbf{Block Flipping}} & \multicolumn{2}{c}{\textbf{Bi-Manual Moving}}\\ 
    \textbf{ } & Spatial & Object & Spatial & Object\\ 
    \toprule
    3D FM &  0/27 &  0/12 & 6/10 & 3/9\\ 
    \rowcolor{gray!7} 3D FM Comp &  0/27 &  0/12 & \textbf{8/10} & \textbf{7/9} \\ 
    3D FM Force &  16/27 &  4/12 & 7/10 & 6/9\\ 
    \rowcolor{gray!7} \textbf{3D FM Comp + Force} &  \textbf{24/27} & \textbf{10/12}  & \textbf{8/10} & \textbf{7/9}\\ 
    \bottomrule
    \end{tabular}
    \end{minipage}
\vspace{-10pt}
\end{table*}

\noindent Our experiments aim to answer the following questions:
\begin{itemize}
\item What data can our force-informed data generation scheme offer for learning visuo-force policies?
\item How well can the policy adapt to the real world with generated data (in terms of spatial and object generalization and matching the desired profile)?
\item How does the passive impedance controller perform compared to the classical impedance controller for rolling out the adaptive compliant visuo-force policy?
\end{itemize}

We evaluate our pipeline on two real-world tasks: \textbf{Block Flipping} and \textbf{Bi-manual Moving}. 
In \textbf{Block Flipping}, a single arm pivots a block from flat to upright (\(\approx85^\circ\)) and must keep it stable. 
In \textbf{Bi-manual Moving}, two arms cooperatively place a large object onto a goal platform in a centralized manner; a trial succeeds when the object is stably stacked. Data are collected in IsaacGym using FACTR \cite{liu2025factr} for teleoperation; real-world runs use Franka arms. 
End-effector force is inferred from Franka joint torques, and perception uses an Intel RealSense L515 LiDAR. To reduce the Sim2Real gap for the point cloud data, we inject noise into simulated point clouds, such as vertical “flying pixels” and occlusion shadows by detecting depth jumps and interpolating across them. We also normalize simulated force, which the real forces are also normalized during real deployment.

\subsection{Generating Force-Informed and Point Cloud Data}

To evaluate our simulation-based data generation method, we conduct experiments on \textbf{Block Flipping} and 
\textbf{Bi-manual Moving} with domain randomization over object pose, mass, friction, and end-effector pose  (Table~\ref{tab:randomization}). We intentionally omit object shape variations to test whether simple simulation geometry, combined with our proposed components, suffices for robust real-world deployment.

For \textbf{Block Flipping}, 303 trajectories are generated for policy learning. During the data generation, the robot follows the reference trajectory as shown in (Fig.~\ref{fig:pipeline}) , switches to object-centric replay upon contact, and tracks force-informed virtual targets with a forward speed to maintain contact and complete the flipping stroke. This achieves a success rate of \(97.6\%\). For \textbf{Bi-manual Moving}, 210 trajectories are generated. Both arms follow the same pipeline, using object-centric replay and force-informed targets to maintain coordinated grasps. Quaternion interpolation with SLERP ensures smooth reorientation even under box translation and rotation, yielding robust \(SE(3)\)-adaptive behavior with \(82.9\%\) success. In both tasks, removing Laplacian editing or force-informed targets reduces the success rate to \(0\%\).

\subsection{Deploying the Compliant Policy in the Real World}
Given the generated simulation data, the policy is trained with 3000 epochs. The inference machine uses an RTX 3090 Ti GPU. We evaluated the trained policy in real-world settings to demonstrate spatial and object generalization. We compare to ablation baselines to show that our design is superior for such tasks. While works like \cite{acp} and \cite{chen2025dexforce} have demonstrated the importance of force-informed virtual target, their tasks are with RGB data, naturally lacking 3D understanding. Since our work utilizes point clouds, we ask: would force-informed data still be necessary, given that we have access to the geometric features? How important is force input in these tasks? To answer these questions we provide three baselines: (1) \textbf{3D FM}: A point cloud-only flow matching adapted directly from \cite{Ze2024DP3}, (2) \textbf{3D FM Comp}: A point cloud-only flow matching with compliant output, (3) \textbf{3D FM Force}: Using point cloud and force as input without compliant output, and (4) \textbf{3D FM Comp + Force (\textit{Ours})}: Our full approach. Success rates reported in Table \ref{tab:success_rate}.

\begin{figure}[!t]
    \centering
    \includegraphics[width=0.49\linewidth]{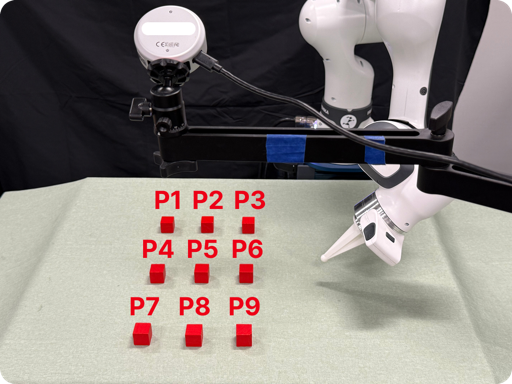}    
    \includegraphics[width=0.49\linewidth]{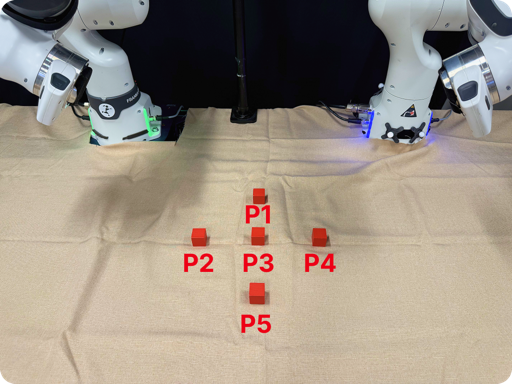}    
    \vspace{-15pt}
    \caption{The red cubes show the location for spatial performance evaluation on (left) \textbf{Block Flipping}  and (right) \textbf{Bi-manual Moving}.  \label{fig:block_spatial}}
    \vspace{-10pt}
\end{figure}

\subsubsection{Block Flipping}

\begin{figure}[!t]
    \centering
    \subfigure[Objects for \textit{Block Flipping} in Real World]{
        \includegraphics[width=0.45\textwidth]{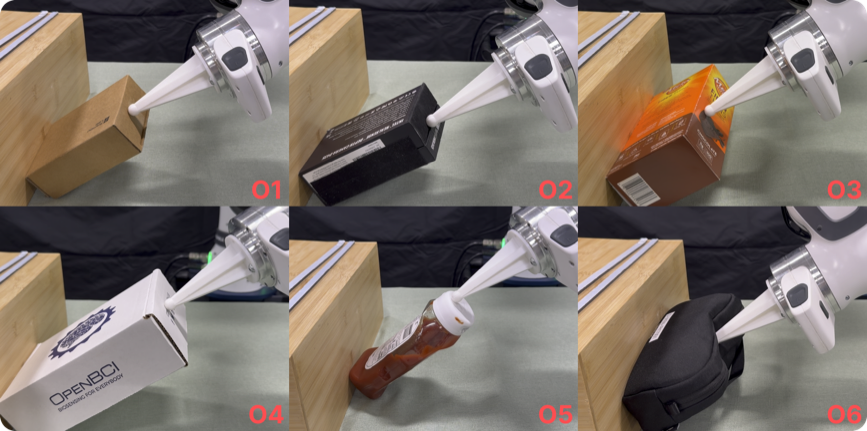}        
        \label{fig:generalization_visual_flip}
    }
    \vspace{-5pt}
    \subfigure[Objects for \textit{Bi-manual Moving} in Real World]{
        \includegraphics[width=0.45\textwidth]{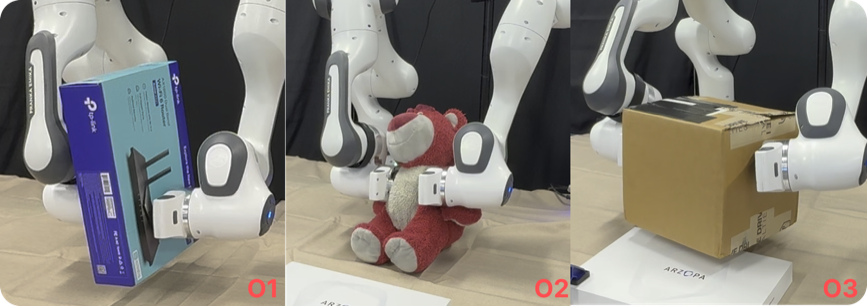}
        \label{fig:generalization_visual_bi}
    }    
    \caption{Real-world object and spatial generalization results.}
    \vspace{-15pt}
\end{figure}

We evaluate real-world spatial generalization at nine testing positions (Fig.~\ref{fig:block_spatial}), as illustrated in Fig.~\ref{fig:block_spatial}. Each numbered position represents a testing site where the center of the test object (Object O1, shown in Fig.~\ref{fig:generalization_visual_flip}) is placed. At each location, we run three trials per method (totaling 27 trials). \textbf{3D FM} and \textbf{3D FM Comp} had a 0\% success rate, primarily failing due to difficulties in establishing initial contact or getting stuck due to excessive force. \textbf{3D FM Force} succeeded consistently in central positions (P4, P5, P6). However, its performance notably degrades on the sides, getting stuck during flipping due to rigid interactions. Our method achieved consistent success at all locations except P7, likely due to limited training coverage or kinematic constraints. This evaluation shows that our method can generalize spatially in the real world by using one demonstration in the simulation.


One might question that using a simple box geometry in generating point cloud data won't generalize across objects in the real world. We use only one size box geometry to generate data and train the policy. We then test it in the real world across objects as shown in Fig. \ref{fig:generalization_visual_flip}. We tested all objects at P5 with two different seeds. The last column of Table \ref{tab:success_rate} shows the success rate. Again, \textbf{3D FM} and \textbf{3D FM Comp} fail for all trials. The failure modes are similar to the spatial evaluation. \textbf{3D FM Force} succeeds at O1 and O6 twice, but fails at others, either not making contact (O3, O5) or getting stuck (O2, O4). Our method fails at O5 once, which cannot make contact, and O3 once, which slips back. This shows that even training with simple geometry, point cloud and force could be combined with the adaptive compliant output to allow for better execution.

\begin{figure}[!tbp]
    \centering

    \includegraphics[width=0.40\textwidth]{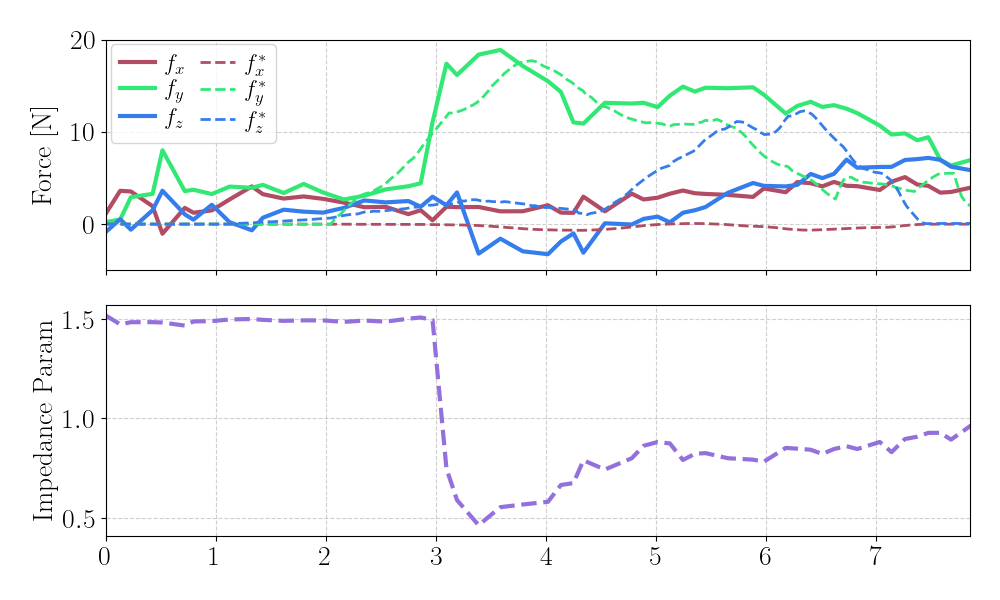}
   
    \vspace{-10pt}
    \caption{Force and impedance profiles from \textbf{Block Flipping }show similar trends between simulation and real. The solid lines are the scaled and smoothed real measurements, while the dashed lines are the force from the simulation demonstration. }
    \label{fig:force_profile_block_flipping}
    \vspace{-5pt}
\end{figure}

Fig. \ref{fig:force_profile_block_flipping} shows the force and compliance profiles comparing simulation data and real-world rollout on one example run. 
The real-world force profiles are scaled to match simulation magnitude for visualization purposes. The plots show that the learned policy deployed in the real world can match the desired behavior in simulation. The $f_y$ increases upon making the first contact and then decreases, while $f_z$ starts to increase. The bottom plot shows the change of the impedance parameter $d$ over time. Upon making the first contact, $d$ decreases to become more compliant and then slowly increases afterwards, matching our design.

\subsubsection{Bi-manual Moving}
We first evaluate real-world spatial generalization at five distinct locations, as shown in Fig. \ref{fig:block_spatial}, using the O3 object in Fig. \ref{fig:generalization_visual_bi}, with each location having two runs. At each run, the object orientation is perturbed slightly to provide more variations. The success rate is shown in Table \ref{tab:success_rate}. The baseline \textbf{3D FM} performs well in the spatial task, with failures happening at P1 and P2. At P1, the object is too close, and both arms tend to move towards the center and quickly forward without the idea of touching the object. Also, \textbf{3D FM} does not perform well on objects O1 and O2, whose sizes are smaller than the training point cloud. On the other hand, \textbf{3D FM Comp} performs well. The outputting virtual target provides a closer grasping distance between the two arms, creating a bigger chance of locking the object. The failures are commonly on the two sides (P2 and P4) with object O3, where the robot cannot avoid the edge of the box when coming down from above to grasp the side. They tend to get stuck on top of the box. Our method and \textbf{3D FM Force} also have the same failure source. The learned policy exhibits recovering behavior from failures. Even when the object drops before the goal platform, the policy recovers by grasping the object again and moving forward until it is on the platform. Such an emergent recovering behavior makes the policy robust in the real world. The goal platform is at a location where the robots can easily get close to singularity in order to place the object on it, so such a readjustment after failure improves the success rate a lot. For different objects, we perform tests with three runs for each object at P5. All methods have a failure with object O1. This object is difficult to recover from and can easily lose control due to its height. \textbf{3D FM Force} fails once with O3, which moves forward without touching the box to obtain the force signal. Force and impedance profiles for this task are in the video. 
\subsection{Passive Impedance for Real World}

We evaluate the effectiveness of passive impedance control in deploying visuomotor policies trained exclusively in simulation. Due to discrepancies and noise inherent in sim2real transfer, policies often exhibit inaccuracies when executed in the real world. We conducted a comparative study between a classical position-based adaptive impedance controller and our proposed passive impedance controller. We tested both controllers using the same trained checkpoint across six distinct objects in a real-world block-flipping task. Each object was consistently placed at P5, and each controller was executed twice per object, resulting in 12 trials.

Our results indicate that passive impedance control achieves a notably higher success rate compared to the classical impedance approach. Specifically, the classical controller encountered two failures with O4 in Fig. \ref{fig:generalization_visual_flip}, which is an object significantly larger than those encountered during training. For these OOD cases, the policy's reference trajectory penetrated the object's geometry. The classical controller, tracking position setpoints, first pushes it hard to attempt to achieve these unreachable positions, then starts to lower the stiffness and progresses to the lift-up stage. In contrast, the passive impedance controller, leveraging velocity-based control, naturally yielded without enforcing strict positional constraints, successfully adapting and completing the task. Additionally, we measured the total energy 
injected into the environment during successful trials in the period of making contact. We show that passive impedance control injects less energy on average over different objects, highlighting its compliance during real-world execution.

\begin{table}[!tbp]
\centering
\caption{Compliant Controller Comparison \label{tab:force_inform}}
\begin{tabular}{cccc}
\toprule
Controller Type & \textbf{Success Rate} & \textbf{Energy Mean (J)} \\
\hline
Classical Impedance & 8/12 & 0.835\\
\rowcolor{gray!7}  \textbf{Passive Impedance} & \textbf{10/12} & \textbf{0.734} \\
\bottomrule
\end{tabular}

\end{table}


\section{DISCUSSION \& CONCLUSION}
We have presented a full framework for (1) generating force and demonstration‑informed simulation data, (2) learning 3D adaptive compliant flow matching policies, and (3) rolling out compliant policies with a passive impedance controller using a state vector field. Our experiments show that the policy can adjust its compliance in response to both point cloud and force inputs, enabling more robust execution in tasks with continuous contact than methods that rely solely on positional accuracy. Zero‑shot deployment on real Franka robots demonstrates that even without real‑world data, the policy executes reliably in both tasks. We also find that the passive compliance rollout reduces energy injection and even improves success. Some limitations remain for future work. For example, creating a simulation environment could require some effort for complicated tasks. Also, our work does not adapt to objects with very different physical properties. Last but not least, our method considers the point cloud only, which cannot achieve tasks that require color information. Overall, our results suggest that a simple sim‑based data generation scheme paired with force‑aware, compliant execution can significantly reduce the dependency on real‑world data while maintaining performance.



\section*{ACKNOWLEDGMENT}
This work was supported by the National Science Foundation (NSF) Foundational Research in Robotics (FRR) program under NSF CAREER Award Grant No. FRR-2443721.


\bibliographystyle{IEEEtran}
\bibliography{root}

\end{document}